\documentclass[10pt,twocolumn,letterpaper]{article}

\usepackage{iccv}
\usepackage{times}
\usepackage{epsfig}
\usepackage{graphicx}
\usepackage{amsmath}
\usepackage{amssymb}

\usepackage[textfont=small,aboveskip=1pt,belowskip=2pt]{subcaption}
\usepackage[table]{xcolor}
\usepackage{blindtext}
\usepackage{array}
\usepackage{multirow}
\usepackage{booktabs}
\usepackage{siunitx}
\usepackage{dcolumn}
\newcolumntype{L}{D{.}{.}{2,5}}
\usepackage{balance}
\usepackage{arydshln}

\usepackage[pagebackref=true,breaklinks=true,letterpaper=true,colorlinks,bookmarks=false]{hyperref}

\newcommand{\obsval}{\mathbf{n}}
\newcommand{\obsvali}{n}
\newcommand{\obsdensity}{p}
\newcommand{\obsdata}{\mathcal{D}}
\newcommand{\obsnum}{M}

\newcommand{\flowvar}{\mathcal{X}}
\newcommand{\flowval}{\mathbf{x}}

\newcommand{\flowdim}{D}
\newcommand{\flownum}{N}
\newcommand{\flowfunc}{f}
\newcommand{\flowinv}{g}
\newcommand{\flowfuncJ}{\mathbf{J}}

\newcommand{\flowparams}{\theta}

\newcommand{\basevar}{{\flowvar_0}}
\newcommand{\baseval}{{\flowval_0}}

\newcommand{\basedensity}{p_\basevar}

\newcommand{\R}{\mathbb{R}}

\newcommand{\params}{\Theta}
\newcommand{\diag}{\textrm{diag}}

\renewcommand{\eg}{e.g.} 
\renewcommand{\ie}{i.e.}

\newcommand{\comment}[1]{ }

\newcommand{\cleanim}{\mathbf{I}}
\newcommand{\gcleanim}{\cleanim_{\gain}}
\newcommand{\noisyim}{\mathbf{\Tilde{I}}}
\newcommand{\noisevar}{\obsval}
\newcommand{\noiseval}{\obsvali}
\newcommand{\poiscomp}{\alpha}
\newcommand{\gauscomp}{\delta}
\newcommand{\scale}{\mathbf{s}}
\newcommand{\gain}{\gamma}
\newcommand{\camparam}{\psi}
\newcommand{\camidx}{m}

\newcommand{\ISO}{\textrm{ISO}}

\newcommand{\trainset}{\obsdata_\mathrm{r}}
\newcommand{\testset}{\obsdata_\mathrm{s}}
\newcommand{\KL}{D_{KL}}

\newcommand{\nll}{\mathit{NLL}}

\newcommand{\gcl}{\cellcolor{green!25}}

\newcommand{\aarnk}[1]{\gcl\fontseries{b}\selectfont{#1}}

\newcommand{\parag}[1]{\noindent\textbf{#1}~~}
\newcommand{\paragskip}[1]{\medskip\noindent\textbf{#1}~~}

\iccvfinalcopy 

\def\iccvPaperID{33} 

\ificcvfinal\fi
\begin{document}

\title{Noise Flow: Noise Modeling with Conditional Normalizing Flows}
\author{%
    Abdelrahman Abdelhamed$^{1,2}$\\
    $^1$York University
    \and
    Marcus A. Brubaker$^{2}$\\
    $^2$Borealis AI
    \and
    Michael S. Brown$^{1,3}$\\
    $^3$Samsung AI Center, Toronto
    \and%
    {\tt\small \{kamel,mbrown\}@eecs.yorku.ca, marcus.brubaker@borealisai.com}\vspace{-3mm}
}

\maketitle



\begin{abstract}
Modeling and synthesizing image noise is an important aspect in many computer vision applications. The long-standing additive white Gaussian and heteroscedastic (signal-dependent) noise models widely used in the literature provide only a coarse approximation of real sensor noise.  This paper introduces Noise Flow, a powerful and accurate noise model based on recent normalizing flow architectures.   Noise Flow combines well-established basic parametric noise models (e.g., signal-dependent noise) with the flexibility and expressiveness of normalizing flow networks. The result is a single, comprehensive, compact noise model containing fewer than 2500 parameters yet able to represent multiple cameras and gain factors. Noise Flow dramatically outperforms existing noise models, with 0.42 nats/pixel improvement over the camera-calibrated noise level functions, which translates to 52\% improvement in the likelihood of sampled noise. Noise Flow represents the first serious attempt to go beyond simple parametric models to one that leverages the power of deep learning and data-driven noise distributions.
\end{abstract}


\section{Introduction}

Image noise modeling, estimation, and reduction is an important and active research area (\eg,~\cite{Foi2009ClippedDenoising, Hwang2012Difference-basedDistribution, Seybold2013TowardsNoise, Trussell2012TheCameras}) with a long-standing history in computer vision (\eg,~\cite{Healey1994RadiometricEstimation, Kuan1985AdaptiveNoise, Liu2008AutomaticImage, Naderi1978EstimationNoise}).  A primary goal of such efforts is to remove or correct for noise in an image, either for aesthetic purposes, or to help improve other downstream tasks.
Towards this end, accurately modeling noise distributions is a critical step.

Existing noise models are not sufficient to represent the complexity of real noise~\cite{Abdelhamed2018ACameras, Plotz2017BenchmarkingPhotographs}.
For example, a univariate homoscedastic Gaussian model does not represent the fact that photon noise is signal-dependent---that is, the variance of the noise is proportional to the magnitude of the signal.   In turn, the signal-dependent heteroscedastic model~\cite{Foi2009ClippedDenoising, Foi2015PracticalRaw-data, Makitalo2013OptimalNoise}, often referred to as the noise level function (NLF), does not represent the spatial non-uniformity of noise power (\eg, fixed-pattern noise) or other sources of noise and non-linearities, such as amplification noise and quantization~\cite{Holst1998CCDDISPLAYS}.
See Figure \ref{fig:overview}.
In spite of their well-known limitations, these models are still the most commonly used.   More complex models, such as a Poisson mixture~\cite{Jin2013ApproximationsNoise, Zhang2017ImprovedNoise}, exist, but still do not capture the complex noise sources mentioned earlier. 

\begin{figure}[!t]
\centering
\includegraphics[width=\linewidth]{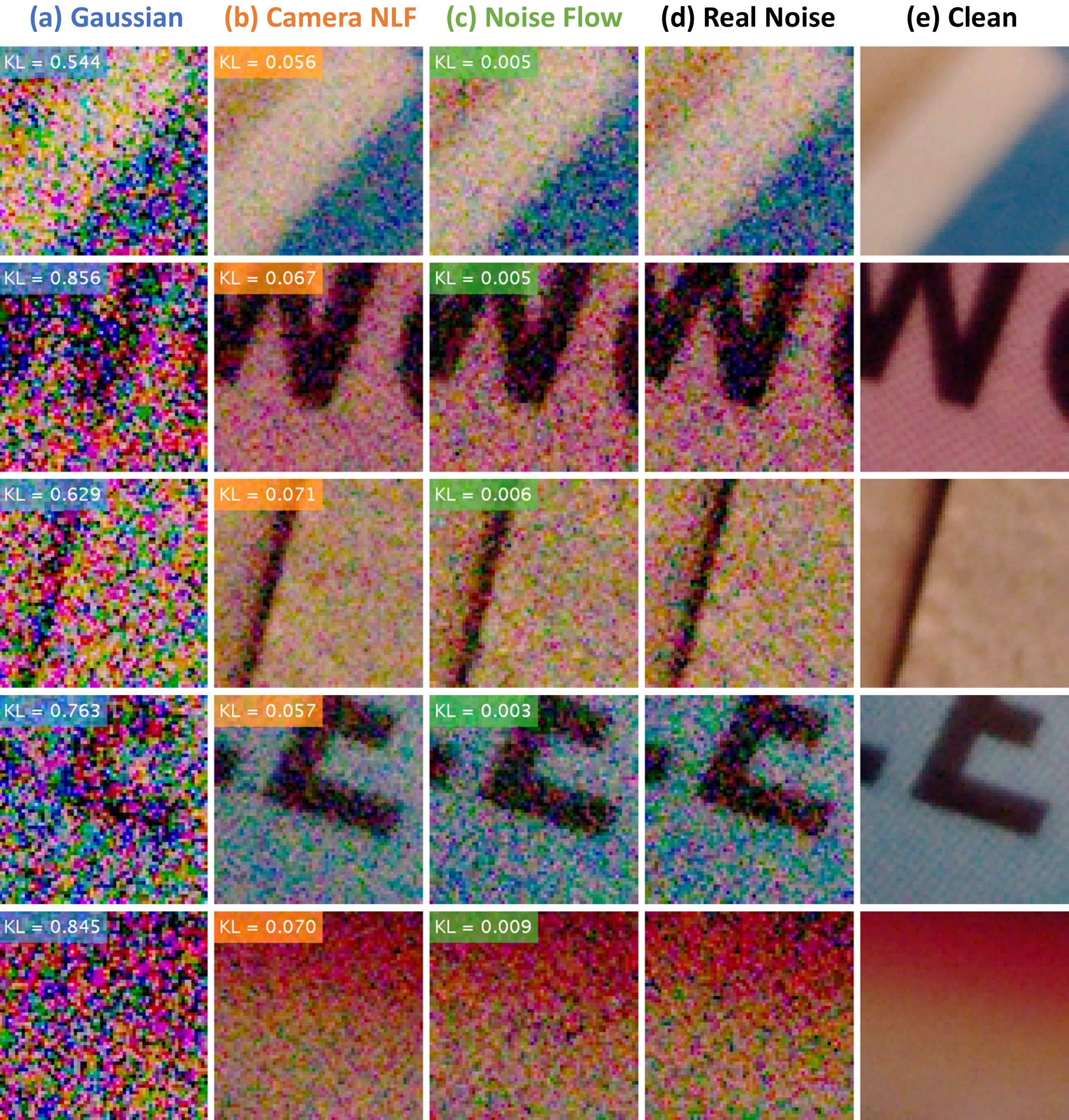}
\caption{\label{fig:teaser} Synthetic noisy images generated by (a) a Gaussian model, (b) a heteroscedastic signal-dependent model represented by camera noise level functions (NLF), and (c) our Noise Flow model. Synthetic noise generated from Noise Flow is consistently the most similar to the real noise in (d), qualitatively and quantitatively (in terms of KL divergence relative to the real noise, shown on each image). (e) Reference clean image. Images are from the SIDD~\protect\cite{Abdelhamed2018ACameras}.%
\vspace{-3mm}%
}
\end{figure}


\paragskip{Contribution}~We introduce Noise Flow, a new noise model that combines the insights of parametric noise models and the expressiveness of powerful generative models.
Specifically, we leverage recent normalizing flow architectures~\cite{Kingma2018Glow:Convolutions} to accurately model noise distributions observed from large datasets of real noisy images. 
In particular, based on the recent Glow architecture \cite{Kingma2018Glow:Convolutions}, we construct a normalizing flow model which is conditioned on critical variables, such as intensity, camera type, and gain settings (\ie, ISO).
The model can be shown to be a strict generalization of the camera NLF but with the ability to capture significantly more complex behaviour.
The result is a single model that is compact (fewer then 2500 parameters) and considerably more accurate than existing models. See Figure~\ref{fig:teaser}. 
We explore different aspects of the model through a set of ablation studies.  To demonstrate the effectiveness of Noise Flow, we consider the application of denoising and use Noise Flow to synthesize training data for a denoising CNN resulting in significant improvements in PSNR.
Code and pre-trained models for Noise Flow are available at: \url{https://github.com/BorealisAI/noise_flow}.


\section{Background and Related Work}

Image noise is an undesirable by-product of any imaging system. Image noise can be described as deviations of the measurements from the actual signal and results from a number of causes, including physical phenomena, such as photon noise, or the electronic characteristics of the imaging sensors, such as fixed pattern noise. 

Given an observed image $\noisyim$ and its underlying noise-free image $\cleanim$, their relationship can be written as
\begin{equation}
    \noisyim = \cleanim + \noisevar,
\label{eq:noise}
\end{equation}
where $\noisevar$ is the noise corrupting $\cleanim$. Our focus in this work is to model $\noisevar$.

Several noise models have been proposed in the literature. The simplest and most common noise model is the homoscedastic Gaussian assumption, also known as the additive white Gaussian noise (AWGN). Under this assumption, the distribution of noise in an image is a Gaussian distribution with independent and identically distributed values:
\begin{equation}
    \noiseval_i \sim \mathcal{N}(0, \sigma^2),
\label{eq:AWGN}
\end{equation}
where $\noiseval_i$ is the noise value at pixel $i$ and follows a normal distribution with zero mean and $\sigma^2$ variance.

Despite its prevalence, the Gaussian model does not represent the fact that photon noise is signal-dependent. To account for signal dependency of noise, a Poisson distribution $\mathcal{P}$ is used instead:
\begin{equation}
\noiseval_i \sim \alpha \mathcal{P}(\cleanim_i) - \cleanim_i,
\end{equation}
where $\cleanim_i$, the underlying noise-free signal at pixel $i$, is both the mean and variance of the noise, and $\alpha$ is a sensor-specific scaling factor of the signal.

Neither the Gaussian nor the Poisson models alone can accurately describe image noise. That is because image noise consists of both signal-dependent and signal-independent components. To address such limitation, a Poisson-Gaussian model has been adapted~\cite{Foi2009ClippedDenoising, Foi2015PracticalRaw-data, Makitalo2013OptimalNoise}, where the noise is a combination of a signal-dependent Poisson distribution and a signal-independent Gaussian distribution:
\begin{equation}
    \noiseval_i \sim \poiscomp \ \mathcal{P}(\cleanim_i) - \cleanim_i + \mathcal{N}(0, \gauscomp^2).
\end{equation}

A more widely accepted alternative to the Poisson-Gaussian model is to replace the Poisson component by a Gaussian distribution whose variance is signal-dependent~\cite{Liu2014PracticalImage, Mohsen1975NoiseDevices}, which is referred to as the heteroscedastic Gaussian model:
\begin{equation}
\noiseval_i  \sim \mathcal{N}(0, \poiscomp^2 \ \cleanim_i + \gauscomp^2).
\label{eq:sdn}
\end{equation}
The heteroscedastic Gaussian model is more commonly referred to as the noise level function (NLF) and describes the relationship between image intensity and noise variance:
\begin{equation}
    \mathrm{var}(\noiseval_i) = \beta_1 \ \cleanim_i + \beta_2, \qquad \beta_1 = \poiscomp^2, \beta_2 = \gauscomp^2.
\label{eq:sdn-var}
\end{equation}

Signal-dependent models may accurately describe noise components, such as photon noise. However, in real images there are still other noise sources that may not be accurately represented by such models~\cite{Abdelhamed2018ACameras, Foi2009ClippedDenoising,Plotz2017BenchmarkingPhotographs}. Examples of such sources include fixed-pattern noise, defective pixels, clipped intensities, spatially correlated noise (\ie, cross-talk), amplification, and quantization noise. Some attempts have been made to close the gap between the prior models and the realistic cases of noise---for example, using a clipped heteroscedastic distribution to account for clipped image intensities~\cite{Foi2009ClippedDenoising} or using a Poisson mixture model to account for the tail behaviour of real sensor noise~\cite{Zhang2017ImprovedNoise}.
Recently, a GAN was trained for synthesizing noise~\cite{Chen2018ImageModeling}; however, it was not clear how to quantitatively assess the quality of  the generated samples.
To this end, there is still a lack of noise models that capture the characteristics of real noise. 
In this paper, we propose a data-driven normalizing flow model that can estimate the density of a real noise distribution. Unlike prior attempts, our model can capture the complex characteristics of noise that cannot be explicitly parameterized by existing models.

\subsection{Normalizing Flows}

Normalizing flows were first introduced to machine learning in the context of variational inference \cite{Rezende2015VariationalFlows} and density estimation \cite{Dinh2015NICE:Estimation} and are seeing increased interest for generative modeling \cite{Kingma2018Glow:Convolutions}.
A normalizing flow is a transformation of a random variable with a known distribution (typically Normal) through a sequence of differentiable, invertible mappings.
Formally, let $\baseval \in \R^\flowdim$ be a random variable with a known and tractable probability density function $\basedensity : \R^\flowdim \rightarrow \R$ and let $\flowval_1, \dots, \flowval_\flownum$ be a sequence of random variables such that $\flowval_i = \flowfunc_i(\flowval_{i-1})$ where $\flowfunc_i : \R^\flowdim \rightarrow \R^\flowdim$ is a differentiable, bijective function.
Then if $\obsval = \flowfunc(\baseval) = \flowfunc_\flownum \circ \flowfunc_{\flownum-1} \circ \dots \circ \flowfunc_1(\baseval)$, the change of variables formula says that the probability density function for $\obsval$ is
\begin{equation}
\obsdensity(\obsval) 
= \basedensity(\flowinv(\obsval)) \prod_{j=1}^\flownum \left| \det \flowfuncJ_j (\flowinv(\obsval)) \right|^{-1}
\label{eq:nrm-flow}
\end{equation}
where $\flowinv = \flowinv_1 \circ \dots \circ \flowinv_{\flownum-1} \circ \flowinv_\flownum$ is the inverse of $\flowfunc$, and  $\flowfuncJ_j = \partial \flowfunc_j / \partial \flowval_{j-1}$ is the Jacobian of the $j$th transformation $\flowfunc_j$ with respect to its input $\flowval_{j-1}$ (\ie, the output of $\flowfunc_{j-1}$).

\paragskip{Density Estimation}
A normalizing flow can be directly used for density estimation by finding parameters which maximize the log likelihood of a set of samples.
Given the observed data, $\obsdata = \{ \obsval_i \}_{i=1}^{\obsnum}$, and assuming the transformations $\flowfunc_1,\dots,\flowfunc_\flownum$ are parameterized by $\params = (\flowparams_1,\dots,\flowparams_\flownum)$ respectively, the log likelihood of the data $\log p(\obsdata | \params)$ is
\begin{equation}
\sum_{i=1}^{\obsnum} \log \basedensity(\flowinv(\obsval_i | \params)) - \sum_{j=1}^{\flownum} \log \left| \det \flowfuncJ_j(\flowinv(\obsval_i | \params), \flowparams_j) \right|    
 \label{eq:density-est}
\end{equation}
where the first term is the log likelihood of the sample under the base measure and the second term, sometimes called the log-determinant or volume correction, accounts for the change of volume induced by the transformation by the normalizing flows.

\paragskip{Bijective Transformations}
To construct an efficient normalizing flow we need to define differentiable and bijective transformations $\flowfunc$.
Beyond being able to define and compute $\flowfunc$, we also need to be able to efficiently compute its inverse, $\flowinv$, and the log determinant $\log \left| \det \flowfuncJ \right|$, which are necessary to evaluate the data log likelihood in Equation \ref{eq:density-est}.
First consider the case of a linear transformations~\cite{Kingma2018Glow:Convolutions} 
\begin{equation}
\flowfunc(\flowval) = \mathbf{A} \flowval + \mathbf{b}
\label{eq:linear}
\end{equation}
where $\mathbf{A}\in \R^{\flowdim \times \flowdim}$ and $\mathbf{b} \in \R^\flowdim$ are parameters.
For $\flowfunc$ to be invertible $\mathbf{A}$ must have full rank; its inverse is given by $\flowinv(\flowval) = \mathbf{A}^{-1}(\flowval - \mathbf{b})$ and the determinant of the Jacobian is simply $\det \flowfuncJ = \det \mathbf{A}$.


\paragskip{Affine Coupling}
To enable more expressive transformations, we can use the concept of coupling~\cite{Dinh2015NICE:Estimation}.
Let $\flowval = (\flowval_A,\flowval_B)$ be a disjoint partition of the dimensions of $\flowval$ and let $\hat{\flowfunc}(\flowval_A | \theta)$ be a bijection on $\flowval_A$ which is parameterized by $\theta$.
Then a coupling flow is
\begin{equation}
\flowfunc(\flowval) = (\hat{\flowfunc}(\flowval_A ; \theta(\flowval_B)), \flowval_B)
\label{eq:coupling}
\end{equation}
where $\theta(\flowval_B)$ is \emph{any} arbitrary function which uses only $\flowval_B$ as input.
The power of a coupling flow resides, largely, in the ability of $\theta(\flowval_B)$ to be arbitrarily complex.
For instance, shallow ResNets~\cite{He2016DeepRecognition} were used for this function in~\cite{Kingma2018Glow:Convolutions}.

Inverting a coupling flow can be done by using the inverse of $\hat{\flowfunc}$.
Further, the Jacobian of $\flowfunc$ is a block triangular matrix where the diagonal blocks are $\hat{\flowfuncJ}$ and the identity.
Hence the determinant of the Jacobian is simply the determinant of $\hat{\flowfuncJ}$.
A common form of a coupling layer is the \emph{affine coupling layer}~\cite{Dinh2017DensityNVP,Kingma2018Glow:Convolutions}  
\begin{equation}
\hat{\flowfunc}(\flowval ; \mathbf{a},\mathbf{b}) = \mathbf{D} \flowval + \mathbf{b}
\label{eq:affine-coupling}
\end{equation}
where $\mathbf{D} = \diag(\mathbf{a})$ is a diagonal matrix.
To ensure that $\mathbf{D}$ is invertible and has non-zero diagonals it is common to use $\mathbf{D} = \diag(\exp(\mathbf{a}))$.

With the above formulation of normalizing flows, it becomes clear that we can utilize their expressive power for modeling real image noise distributions and mapping them to easily tractable simpler distributions. As a by-product, such models can directly be used for realistic noise synthesis. Since the introduction of normalizing flows to machine learning, they have been focused towards image generation tasks (\eg,~\cite{Kingma2018Glow:Convolutions}). However, in this work, we adapt normalizing flows to the task of noise modeling and synthesis by introducing two new conditional bijections, which we describe next.

\section{Noise Flow}

In this section, we define a new architecture of normalizing flows for modeling noise which we call Noise Flow.  Noise Flow contains novel bijective transformations which capture the well-established and fundamental aspects of parametric noise models (\eg, signal-dependent noise and gain) which are mixed with more expressive and general affine coupling transformations.

\begin{figure*}
    \centering
    \includegraphics[width=\textwidth]{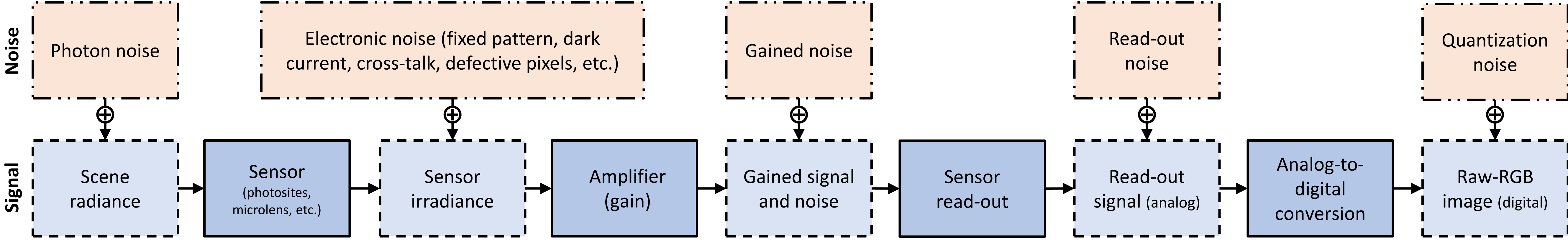}
    \caption{A simplified model of an imaging pipeline showing imaging processes (in the bottom row) and the associated noise processes (in the top row). Model adapted from~\protect\cite{Gow2007APerformance, Hasinoff2010Noise-OptimalPhotography,Healey1994RadiometricEstimation,Liu2008AutomaticImage}.\label{fig:overview}}
\end{figure*}%
\begin{figure*}
    \centering
    \includegraphics[width=\textwidth]{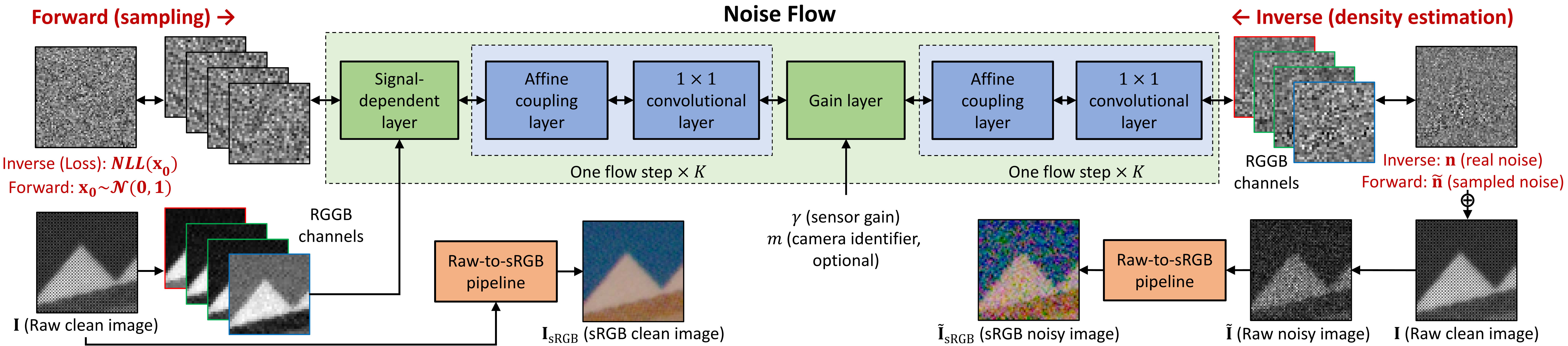}
    \caption{The architecture of our Noise Flow model. The affine coupling and $1\times 1$ convolutional layers are ported from~\protect\cite{Kingma2018Glow:Convolutions}. The signal-dependent and gain layers are newly proposed. The Raw-to-sRGB pipeline is ported from~\protect\cite{Abdelhamed2018ACameras}.\label{fig:noise-flow}\vspace{-3mm}}
\end{figure*}

\subsection{Noise Modeling using Normalizing Flows}

Starting from Equations~\ref{eq:noise} and \ref{eq:density-est}, we can directly use normalizing flows to estimate the probability density of a complex noise distribution.  Let $\obsdata = \{ \obsval_i \}_{i=1}^{\obsnum}$ denote a dataset of observed camera noise where $\obsval_i$ is the noise layer corrupting a raw-RGB image.
Noise layers can be obtained by subtracting a clean image from its corresponding noisy one.
As is common, we choose an isotropic Normal distribution with zero mean and identity covariance as the base measure.
Next, we choose a set of bijective transformations, with a set of parameters $\params$, that define the normalizing flows model.
Lastly, we train the model by minimizing the negative log likelihood of the transformed distribution, as indicated in Equation~\ref{eq:density-est}.

We choose the Glow model~\cite{Kingma2018Glow:Convolutions} as our starting point.
We use two types of bijective transformations (\ie, layers) from the Glow model: (1) the affine coupling layer as defined in Equation~\ref{eq:affine-coupling} that can capture arbitrary correlations between image dimensions (\ie, pixels); and (2) the $1 \times 1$ convolutional layers that are used to capture cross-channel correlations in the input images.

\subsection{Noise Modeling using Conditional Normalizing Flows}

Existing normalizing flows are generally trained in an unsupervised manner using only data samples and without additional information about the data.
In our case, we have some knowledge regarding the noise processes, such as the signal-dependency of noise and the scaling of the noise based on sensor gain. Some of these noise processes are shown in Figure~\ref{fig:overview} along with their associated imaging processes.  Thus, we propose new normalizing flow layers that are conditional on such information. However, many noise processes, such as fixed-pattern noise, cannot be easily specified directly.
To capture these other phenomena we use a combination of affine coupling layers (Equations~\ref{eq:coupling} and~\ref{eq:affine-coupling}) and $1\times 1$ convolutional layers (a form of Equation~\ref{eq:linear}) which were introduced by the Glow model \cite{Kingma2018Glow:Convolutions}.

Figure~\ref{fig:noise-flow} shows the proposed architecture of our noise model (Noise Flow). Noise Flow is a sequence of a signal-dependent layer; $K$ unconditional flow steps; a gain layer; and another set of $K$ unconditional flow steps. Each unconditional flow step is a block of an affine coupling layer followed by a $1\times 1$ convolutional layer. The term $K$ is the number of flow steps to be used in the model. In our experiments, we use $K = 4$, unless otherwise specified. The model is fully bijective---that is, it can operate in both directions, meaning that it can be used for both simulating noise (by sampling from the base measure $\baseval$ and applying the sequence of transformations) or likelihood evaluation (by using the inverse transformation given a noise sample $\noisyim$ to evaluation of Equation~\ref{eq:nrm-flow}).
The Raw-to-sRGB rendering pipeline is imported from~\cite{Abdelhamed2018ACameras}. Next, we discuss the proposed signal-dependent and gain layers in details.

\subsubsection{Signal-Dependent Layer}

We construct a bijective transformation that mimics the signal-dependent noise process defined in Equation~\ref{eq:sdn}.
This layer is defined as
\begin{equation}
    \flowfunc(\flowval) = \scale \odot \flowval, \qquad
    \scale = (\beta_1 \cleanim + \beta_2)^{\frac{1}{2}}.
\end{equation}
The inverse of this layer is given by $\flowinv(\flowval) = \scale^{-1} \odot \flowval$, where $\cleanim$ is the latent clean image, and $\odot$ is point-wise multiplication.
To account for volume change induced by this transformation, we compute the log determinant as
\begin{equation}
    \log \left| \det \flowfuncJ \right| = \sum_{i=1}^{\flowdim} \log(s_i)
\end{equation}
where $s_i$ is the $i$th element of $\scale$ and $\flowdim$ is the dimensionality (\ie, number of pixels and channels) of $\flowval$.
The signal-dependent noise parameters $\beta_1$ and $\beta_2$ should be strictly positive as the standard deviation of noise should be positive and an increasing function of intensity.
Thus, we parameterize them as %
$\beta_1 = \exp(b_1)$ and $\beta_2 = \exp(b_2)$. We initialize the signal-dependent layer to resemble an identity transformation by setting $b_1 = -5.0$ and $b_2 = 0$. This way,  %
$\beta_1 \approx 0$ and $\beta_2 = 1.0$, and hence the initial scale  $\scale \approx 1.0$.

\subsubsection{Gain Layer}

Sensor gain amplifies not only the signal, but also the noise. With common use of higher gain factors in low-light imaging, it becomes essential to explicitly factor the effect of gain in any noise model. Hence, we propose a gain-dependent bijective transformation as a layer of Noise Flow. 
The gain layer is modeled as a scale factor $\gain$ of the corresponding ISO level of the image, and hence the transformation is
\begin{equation}
    \flowfunc(\flowval) = \gain(\ISO) \odot \flowval, \qquad \gain(\ISO) = u({\ISO}) \times \ISO,%
    \label{eq:gain-layer}
\end{equation}
where $u(\ISO) > 0$ allows the gain factors to vary somewhat from the strict scaling dictated by the ISO value.
The inverse transformation is \ $ \flowinv(\flowval) = \gain^{-1}(\ISO) \odot \flowval$, where $u$ is parameterized to be strictly positive and is initialized to $u \approx 1/200$ to account for the typical scale of the ISO values.
Finally, the log determinant of this layer is
\begin{equation}
    \log \left| \det \flowfuncJ \right| = \flowdim \log(\gain(\ISO)),
\end{equation}
where $\flowdim$ is the number of dimensions (\ie, pixels and channels) in $\flowval$.
There are many ways to represent $u(\ISO)$.
However, since the available dataset contained only a small set of discrete ISO levels, we chose to simply use a discrete set of values.
Formally $u(\ISO) = \exp(v_{\ISO})$  where the exponential is used to ensure that $u(\ISO)$ is positive. We use a single parameter for each ISO level in the dataset (\eg, $\{v_{100}, \dots, v_{1600}\}$).
The values of $v_\ISO$ are initialized so that $\exp(v_\ISO)\approx 1/200$ to account for the scale of the ISO value and ensure the initial transformation remains close to an identity transformation.

Different cameras may have different gain factors corresponding to their ISO levels.
These camera-specific gain factors are usually proprietary and hard to access but may have a significant impact on the noise distribution of an image. 
To handle this, we use an additional set of parameters to adjust the gain layer for each camera.
In this case, the above gain layer is adjusted by introducing a camera-specific scaling factor.
That is, 
\begin{equation}
    \gain(\ISO, \camidx) = \camparam_\camidx \times u(\ISO) \times \ISO,
\end{equation}
where $\camparam_\camidx \in \R^+$ is the scaling factor for camera $\camidx$.
This is a simple model but was found to be effective to capture differences in gain factors between cameras.


\section{Experiments}

To assess the performance of Noise Flow, we train it to model the realistic noise distribution of the Smartphone Image Denoising Dataset (SIDD)~\cite{Abdelhamed2018ACameras} and also evaluate the sampling accuracy of the trained model.

\subsection{Experimental Setup}

\parag{Dataset}
We choose the SIDD for training our Noise Flow model. The SIDD consists of thousands of noisy and corresponding ground truth images, from ten different scenes, captured repeatedly with five different smartphone cameras under different lighting conditions and ISO levels. The ISO levels ranged from 50 to 10,000. The images are provided in both Raw-RGB and sRGB color spaces. We believe this  dataset is the best fit to our task for noise modeling, mainly due to the great extent of variety in cameras, ISO levels, and lighting conditions.

\paragskip{Data preparation}%
We start by collecting a large number of realistic noise samples from the SIDD. We obtain the noise layers by subtracting the ground truth images from the noisy ones.  In this work, we use only raw-RGB images as they directly represent the noise distribution of the underlying cameras. We avoid using sRGB images as rendering image into sRGB space tends to significantly change the noise distribution~\cite{Nam2016ADenoising}. 
We arrange the data as approximately $500,000$ image patches of size $64 \times 64$ pixels.
We split the data into a training set $\trainset$ of approximately $70\%$ of the data and a testing set $\testset$ of approximately $30\%$ of the data.
We ensure that the same set of cameras and ISO levels is represented in both the training and testing sets.
For visualization only, we render raw-RGB images  through a color processing pipeline into sRGB color space. 

The SIDD provides only the gain amplified clean image $\gcleanim$ and not the true latent clean image $\cleanim$.
To handle this, we use the learned gain parameter $\gain$ to correct for this and estimate the latent clean image as $\cleanim = \gcleanim / \gain$ when it is needed in the signal-dependant layer.

\paragskip{Loss function and evaluation metrics} 
We train Noise Flow as a density estimator of the noise distribution of the dataset which can be also used to generate noise samples from this distribution. For density estimation training, we use the negative log likelihood ($\nll$) of the training set (see Equation~\ref{eq:density-est}) as the loss function
%
which is optimized using Adam \cite{Kingma2015Adam:Optimization}. 
For evaluation, we consider the same $\nll$ evaluated on the test set.

To provide further insight in the differences between the approaches, we also consider the Kullback-Leibler (KL) divergence of the pixel-wise marginal distributions between generated samples and test set samples.
Such a measure ignores the ability of a model to capture correlations but focuses on a model's ability to capture the most basic characteristics of the distribution.
Specifically, given an image from the test set, we generate a noise sample from the model and compute histograms of the noise values from the test image and the generated noise and report the discrete KL divergence between the histograms.


\paragskip{Baselines}%
We compare the Noise Flow models against two well-established baseline models.
The first is the homoscedastic Gaussian noise model (\ie, AWGN) defined in Equation~\ref{eq:AWGN}.
We prepare this baseline model by estimating the maximum likelihood estimate (MLE) of the noise variance of the training set, assuming a univariate Gaussian distribution.
The second baseline model is the heteroscedastic Gaussian noise model (\ie, NLF), described in Equations~\ref{eq:sdn} and~\ref{eq:sdn-var}, as provided by the camera devices.
The SIDD provides the camera-calibrated NLF for each image.
We use these NLFs as the parameters of the heteroscedastic Gaussian model for each image.
During testing, we compute the $\nll$ of the testing set against both baseline models.

\subsection{Results and Ablation Studies}

\begin{figure}[t!]
    \centering
    \begin{subfigure}[t]{0.49\columnwidth}
    	\centering
    	\includegraphics[width=\columnwidth]{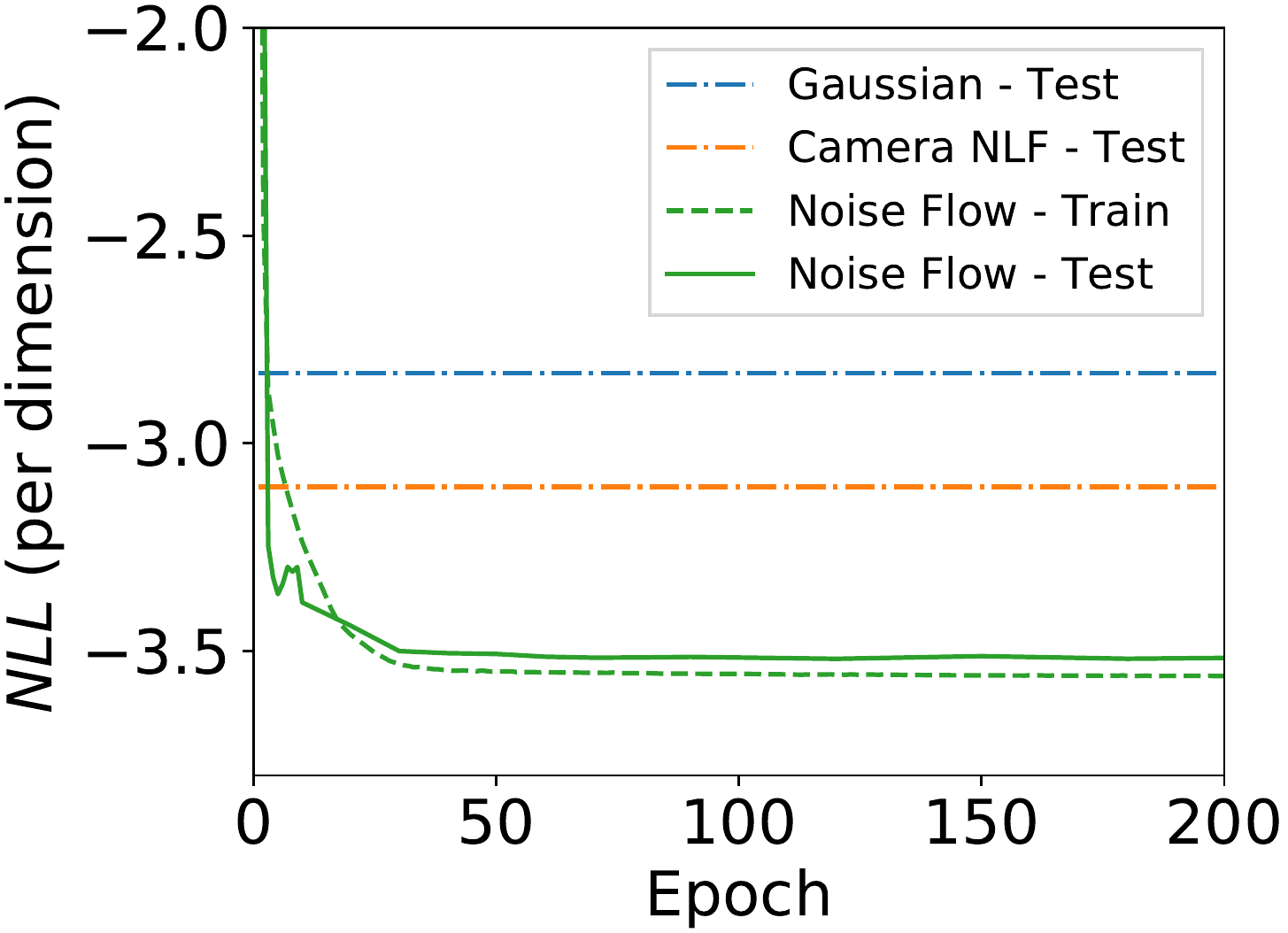}
        \caption{\label{fig:NLL}}
    \end{subfigure}
    \begin{subfigure}[t]{0.49\columnwidth}
    	\centering
    	\includegraphics[width=\columnwidth]{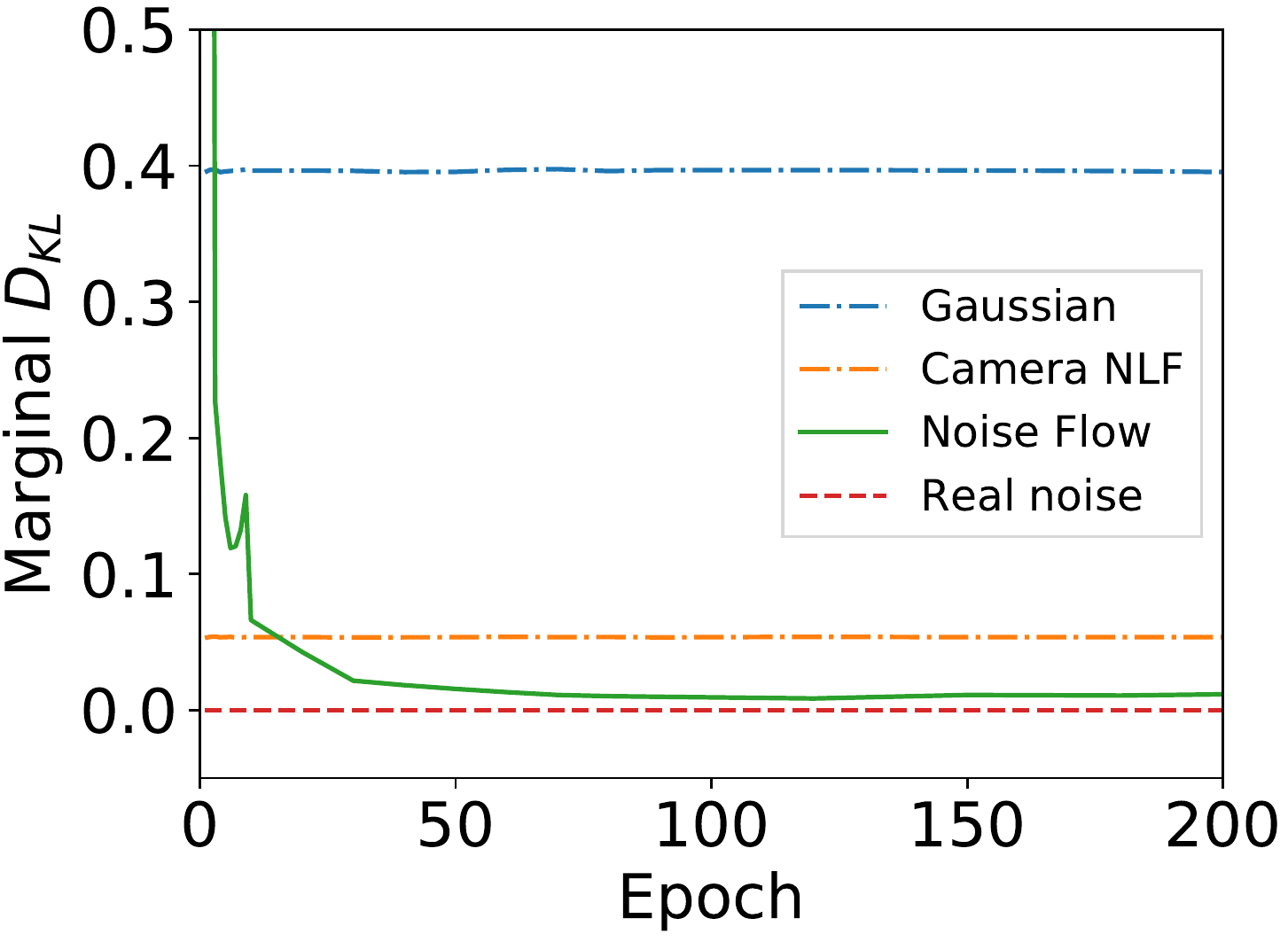}
        \caption{\label{fig:margin-KL}}
    \end{subfigure}
    \caption{(a) $\nll$ per dimension on the training and testing sets of Noise Flow compared to (1) the Gaussian model and (2) the signal-dependent model as represented by the camera-estimated NLFs. (b) Marginal KL divergence ($D_{KL}$) between the generated and the real noise samples.}\vspace{-0mm}
    \label{fig:NLL-KLD}
\end{figure}

\parag{Noise Density Estimation} 
Figure~\ref{fig:NLL} shows the training and testing $\nll$ on the SIDD of Noise Flow compared to (1) the Gaussian noise model and (2) the signal-dependent noise model as represented by the camera-estimated noise level functions (NLFs). It is clear that Noise Flow can model the realistic noise distribution better than Gaussian and signal-dependent models. As shown in Table~\ref{tab:nll-kld}, Noise Flow achieves the best $\nll$,  with $0.69$ and $0.42$ nats/pixel improvement over the Gaussian and camera NLF models, respectively. This translates to $99.4\%$ and $51.6\%$ improvement in likelihood, respectively. We calculate the improvement in likelihood by calculating the corresponding improvement in $\exp(-\nll)$. 

\begin{table}[t]
\centering
\resizebox{\columnwidth}{!}{
\begin{tabular}{lrrc}
\toprule
& \multicolumn{1}{c}{Gaussian} & \multicolumn{1}{c}{Cam. NLF} & \multicolumn{1}{c}{Noise Flow} \\ \midrule
$\nll$ & $-2.831$ ($99.4\%$) & $-3.105$ ($51.6\%$) & \aarnk{\num{-3.521}} \\ 
$D_{KL}$ & $0.394$ ($97.9\%$) & $0.052$ ($84.1\%$) & \phantom{$-$}\aarnk{\num{0.008}} \\ \bottomrule
\end{tabular}
} 
\caption{Best achieved testing $\nll$ and marginal $D_{KL}$ for Noise Flow compared to the Gaussian and Camera NLF baselines. Relative improvements of Noise Flow on other baselines, in terms of \textit{likelihood}, are in parentheses.}\vspace{-1.5mm}
 \label{tab:nll-kld}
\end{table}



\paragskip{Noise Synthesis} %
Figure~\ref{fig:margin-KL} shows the average marginal KL divergence between the generated noise samples and the corresponding noise samples from the testing set for the three models: Gaussian, camera NLF, and Noise Flow. Noise Flow achieves the best KL divergence, with $97.9\%$ and $84.1\%$ improvement over the Gaussian and camera NLF models, respectively, as shown in Table~\ref{tab:nll-kld}.

\begin{figure}[t]
    \centering
    \includegraphics[width=\columnwidth]{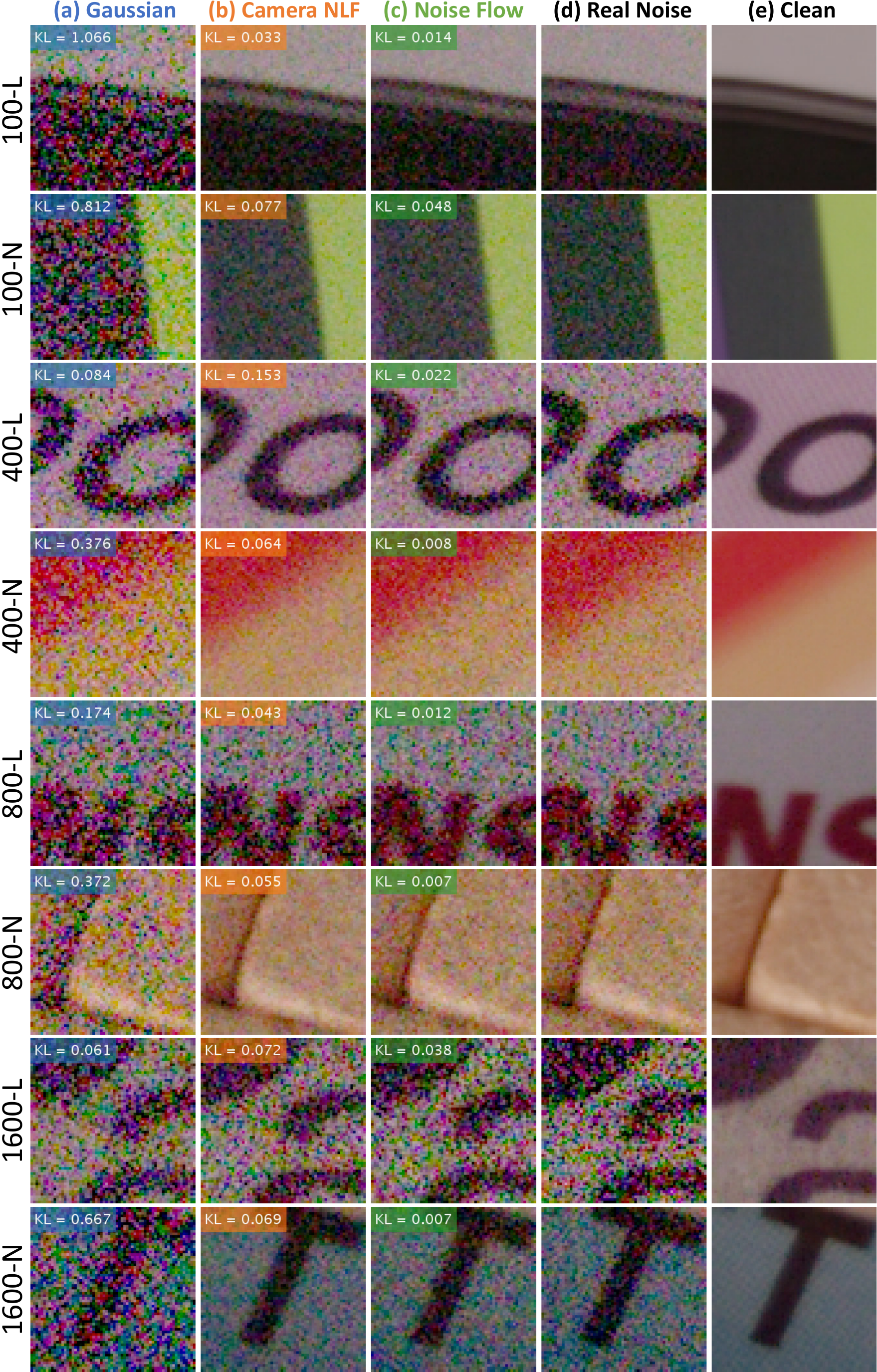}%
    \caption{Generated noise samples from (c) Noise Flow are much closer, in terms of marginal KL divergence, to (d) the real samples; compared to (a) Gaussian and (b) camera NLF models. (e) Clean image. Corresponding ISO levels and lighting conditions are on the left.}%
    \label{fig:noise-samples}%
\end{figure}

Figure~\ref{fig:noise-samples} shows generated noise samples from Noise Flow compared to samples from Gaussian and camera NLF models. We show samples from various ISO levels $\{100, \dots, 1600\}$ and lighting conditions (N: normal light, L: low light). Noise Flow samples are the closest to the real noise distribution in terms of the marginal KL divergence. Also, there are more noticeable visual similarities between Noise Flow samples and the real samples compared to the Gaussian and camera NLF models. 

\begin{figure}[t!]
    \centering
    \begin{subfigure}[t]{0.49\columnwidth}
    	\centering
    	\includegraphics[width=\columnwidth]{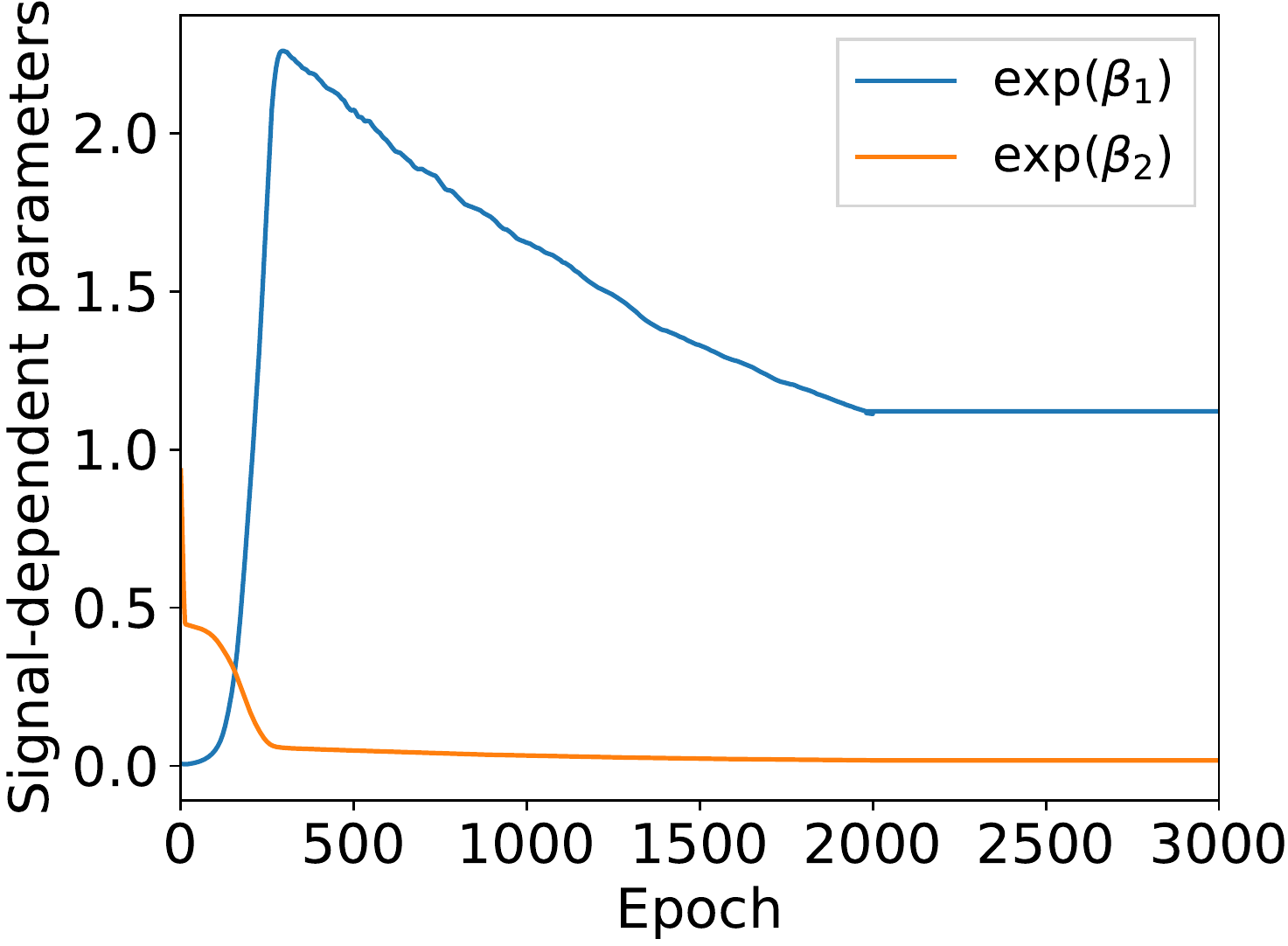}
        \caption{Signal-dependent parameters\label{fig:sdn-params}}
    \end{subfigure}
    \begin{subfigure}[t]{0.49\columnwidth}
    	\centering
    	\includegraphics[width=\columnwidth]{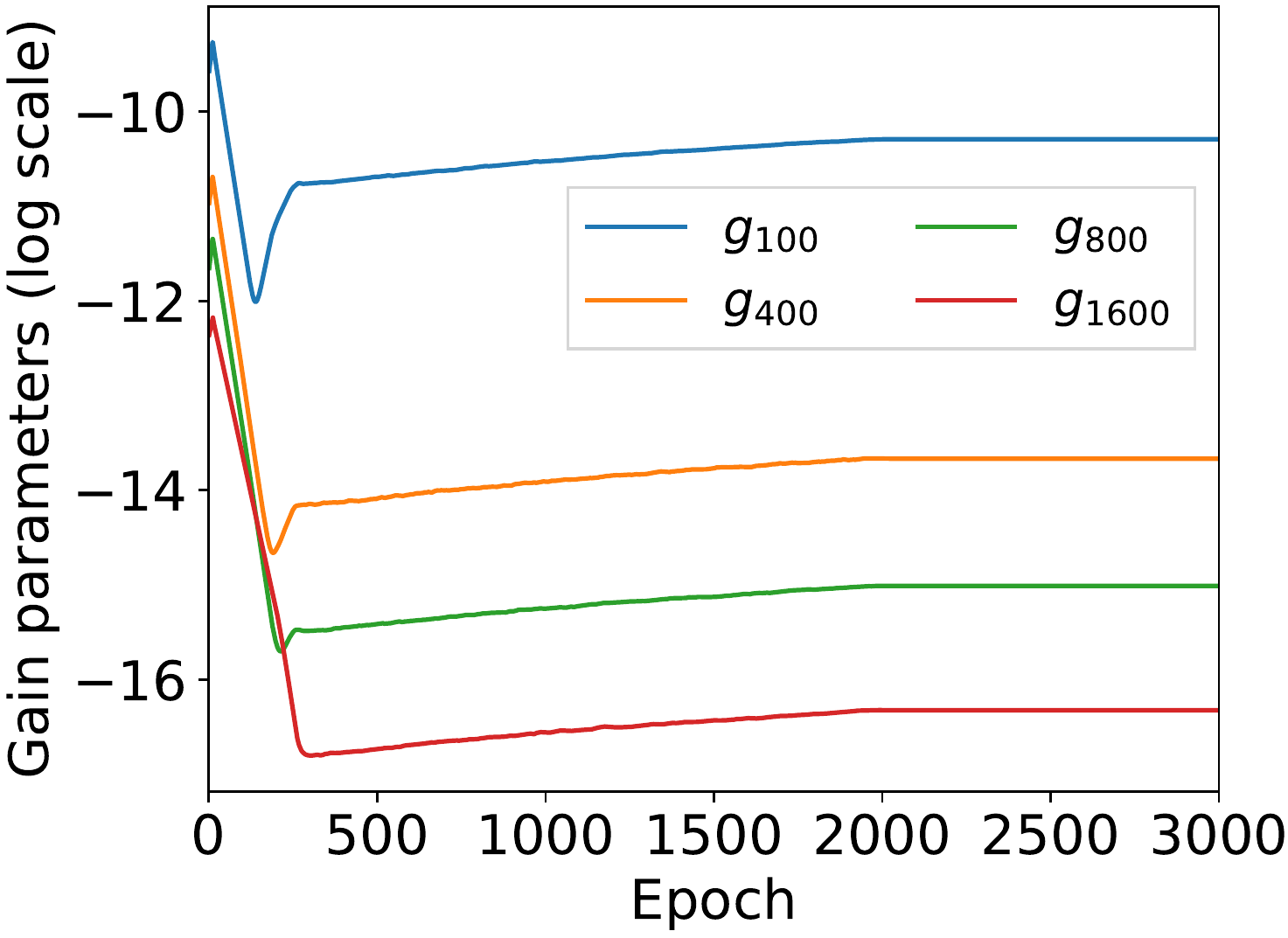}
        \caption{Gain parameters\label{fig:gain-params}}
    \end{subfigure}
    \caption{(a) Signal-dependent noise parameters $\beta_1$ and $\beta_2$ are consistent with the signal-dependent noise model where $\beta_1$ is dominant and $\beta_2$ is much smaller. (b) The gain parameters, in log scale, are consistent with the corresponding ISO levels shown in the legend.\vspace{-3mm}}%
    \label{fig:sdn-and-gain-params}%
\end{figure}%

\paragskip{Learning signal-dependent noise parameters}%
Figure~\ref{fig:sdn-params} shows the learning of the signal-dependent noise parameters $\beta_1$ and $\beta_2$ as defined in Equation~\ref{eq:sdn-var} while training a Noise Flow model. The parameters are converging towards values that are consistent with the signal-dependent noise model where $\beta_1$ is the dominant noise factor that represents the Poisson component of the noise  and $\beta_2$ is the smaller factor representing the additive Gaussian component of the noise. In our experiments, the shown parameters are run through an exponential function to force their values to be strictly positive.

\paragskip{Learning gain factors} %
Figure~\ref{fig:gain-params} shows the learning of the gain factors as defined in Equation~\ref{eq:gain-layer} while training a Noise Flow model. The gain factors $\{\gain_{100}, \dots, \gain_{1600}\}$ are consistent with the corresponding ISO levels indicated by the subscript of each gain factor. This shows the ability of the Noise Flow model to properly factor the sensor gain in the noise modeling and synthesis process. Note that we omitted ISO level 200 from the training and testing sets because there are not enough images from this ISO level in the SIDD.

\begin{figure}[t!]
    \centering
    \begin{subfigure}[b]{0.49\columnwidth}
    	\centering
    	\includegraphics[width=\columnwidth]{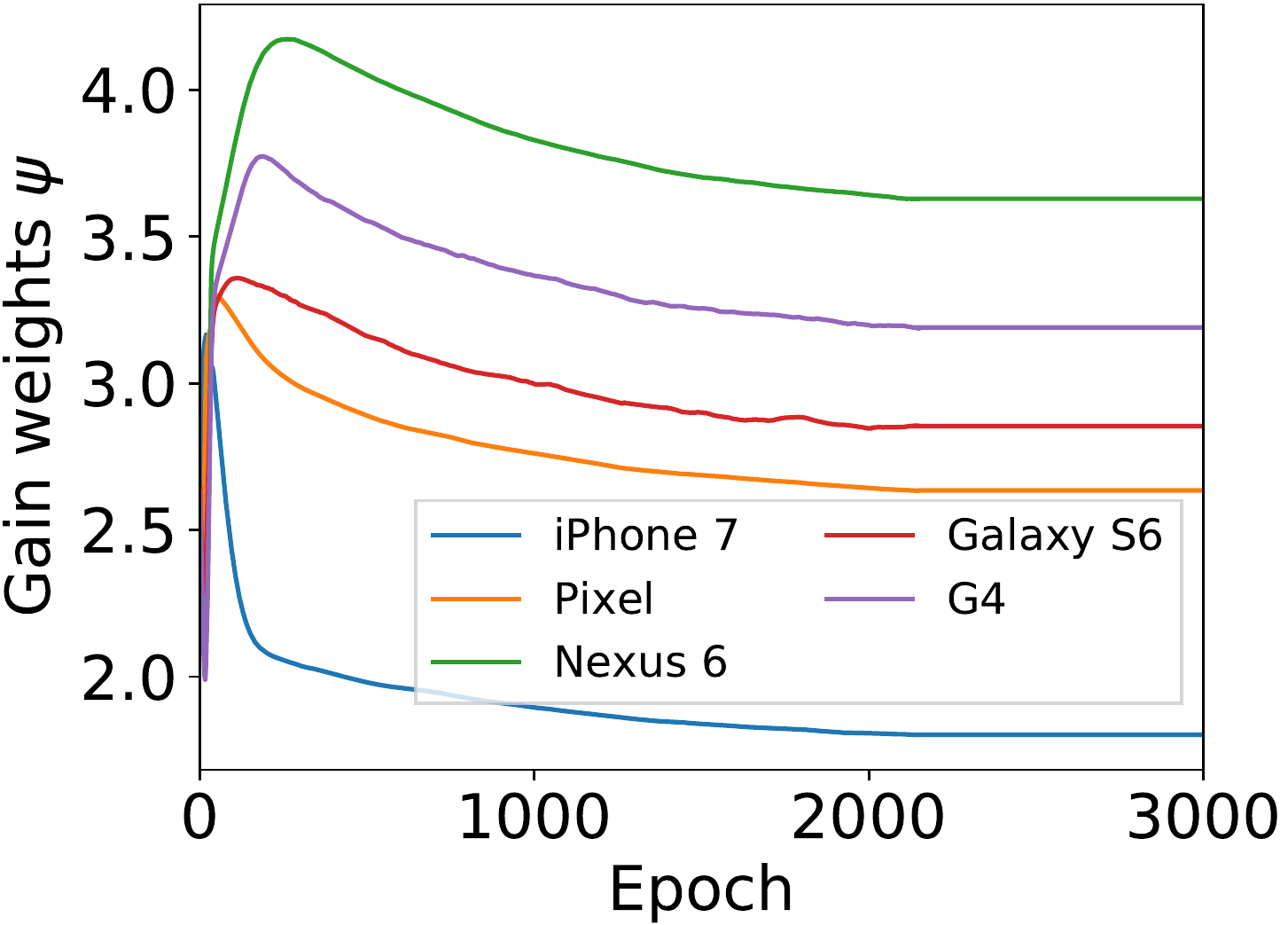}
        \caption{Camera gain weights\label{fig:cam-params-gain}}
    \end{subfigure}%
    \begin{subfigure}[b]{0.49\columnwidth}
	\centering
    	\raisebox{1mm}{\includegraphics[width=\columnwidth]{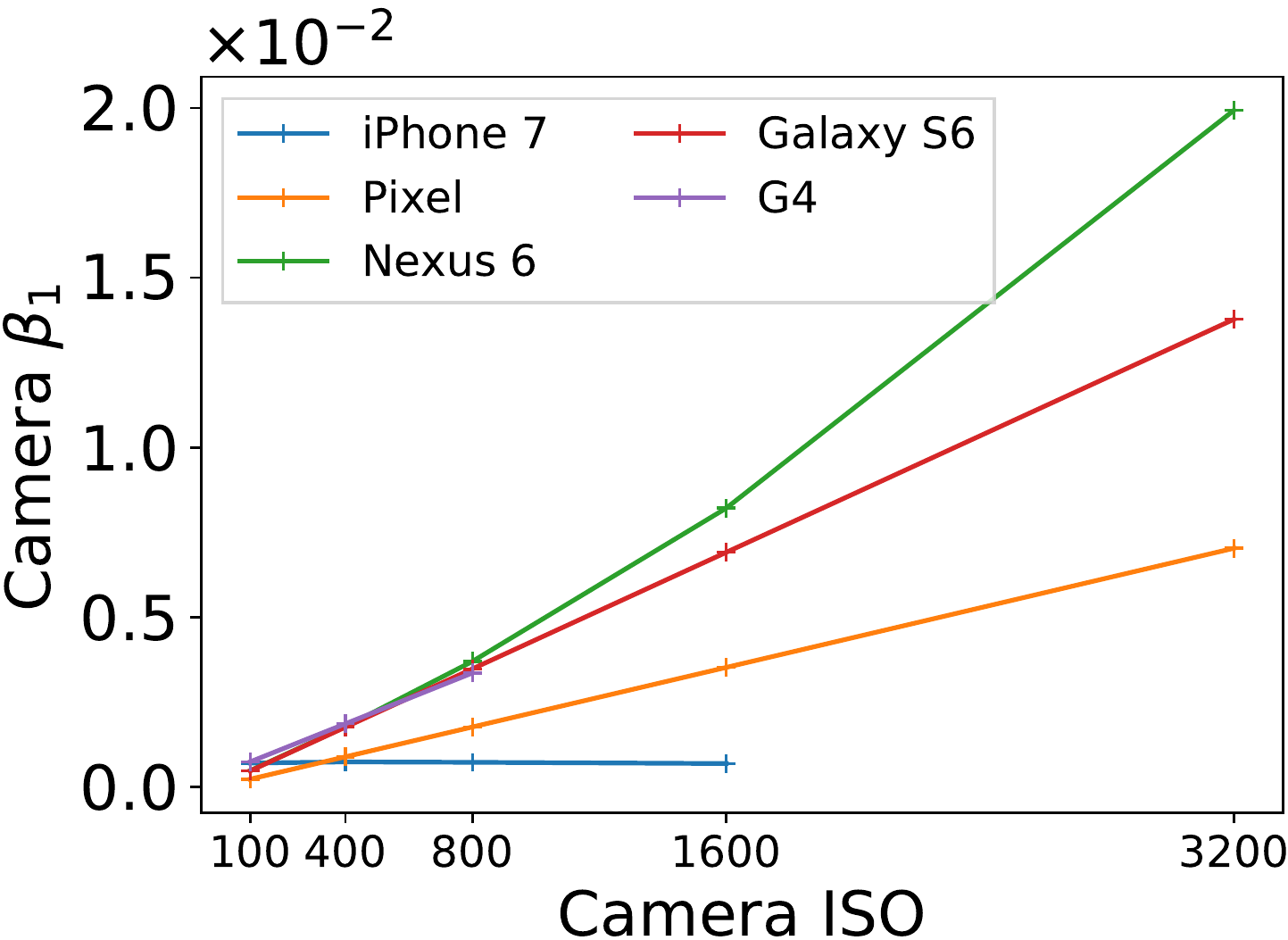}}
        \caption{Camera $\beta_1$ values \label{fig:cam-params-beta1}}
    \end{subfigure}
    \caption{(a) Learned camera-specific weights for the shared gain layer indicates differences in the gain of different cameras. These gains are correlated with the cameras' different values for NLF parameter $\beta_1$ shown in (b). The iPhone and G4 cameras have smaller ranges of ISO values in the SIDD dataset and hence their correlation with the gains is not clear.%
    \vspace{-2mm}}%
    \label{fig:cam-params}%
\end{figure}%

\paragskip{Learning camera-specific parameters} %
In our Noise Flow model, the camera-specific parameters consist of a set of gain scale factors $\{\camparam_\camidx\}$, one for each of the five cameras in the SIDD.
Figure~\ref{fig:cam-params} shows these gain scales for each camera in the dataset during the course of training.
It is clear that there are differences between cameras in the learned gain behaviours.
These differences are consistent with the differences in the noise level function parameter $\beta_1$ of the corresponding cameras shown in Figure~\ref{fig:cam-params-beta1} and capture fundamental differences in the noise behaviour between devices.
This demonstrates the importance of the camera-specific parameters to capture camera-specific noise profiles.
Training Noise Flow for a new camera can be done by fine-tuning the camera-specific parameters within the gain layers; all other layers (i.e., the signal-dependent and affine coupling layers) can be considered non-camera-specific.


\begin{table}[t]
\centering
\resizebox{\columnwidth}{!}{
\begin{tabular}{lll}
\toprule
\multicolumn{1}{c}{Model} & \multicolumn{1}{c}{$\nll$} & \multicolumn{1}{c}{$D_{KL}$} \\ \midrule
S-G & $-3.431$ ($9.42\%$) & $0.067$ ($88.1\%$)\\
S-G-CAM & $-3.511$ ($1.01\%$) & $0.010$ ($20.0\%$)\\ 
S-Ax1-G-Ax1-CAM & $-3.518$ ($0.30\%$) & $0.009$ ($11.1\%$) \\ 
S-Ax4-G-Ax4-CAM & \aarnk{\num{-3.521}} & \aarnk{\num{0.008}} \\ 
(Noise Flow) & & \\ \bottomrule
\end{tabular}
}
\caption{Best achieved testing $\nll$ and marginal $\KL$ for different layer architectures. The symbols S, G, CAM, Ax1, and Ax4 indicate a signal layer, gain layer, camera-specific parameters, one unconditional flow step, and four unconditional flow steps, respectively.   Relative improvements of Noise Flow on each of the other architectures, in terms of likelihood, are in parentheses.\vspace{-3mm}}%
 \label{tab:layer-effect}
\end{table}
%

\paragskip{Effect of individual layers} %
Table~\ref{tab:layer-effect} compares different architecture choices for our Noise Flow model. We denote the different layers as follows: G: gain layer; S: signal-dependent layer; CAM: a layer using camera-specific parameters; Ax1: one unconditional flow step (an affine coupling layer and a $1 \times 1$ convolutional layer); Ax4: four unconditional flow steps.   The results show a significant improvement in noise modeling (in terms of $\nll$ and $D_{KL}$) resulting from the additional camera-specific parameters (\ie, the S-G-CAM model), confirming the differences in noise distributions between cameras and the need for camera-specific noise parameters. Then, we show the effect of using affine coupling layers and $1 \times 1$ convolutional layers in our Noise Flow model.  Adding the Ax1 blocks improves the modeling performance in terms of $\nll$. Also, increasing the number of unconditional flow steps from one to four introduces a slight improvement as well. This indicates the importance of affine coupling layers in capturing additional pixel-correlations that cannot be directly modeled by the signal-dependency or the gain layers. The S-Ax4-G-Ax4-CAM is the final Noise Flow model.

\section{Application to Real Image Denoising}

\parag{Preparation}
To further investigate the accuracy of the Noise Flow model, we use it as a noise generator to train an image denoiser. We use the DnCNN image denoiser~\cite{Zhang2017BeyondDenoising}. We use the clean images from the SIDD-Medium~\cite{Abdelhamed2018ACameras} as training ground truth and the SIDD-Validation as our testing set. The SIDD-Validation contains both real noisy images and the corresponding ground truth. We compare three different cases for training DnCNN using synthetically generated noise: (1) DnCNN-Gauss: homoscedastic Gaussian noise (\ie, AWGN); (2) DnCNN-CamNLF: signal-dependent noise from the camera-calibrated NLFs; and (3) DnCNN-NF: noise generated from our Noise Flow model. For the Gaussian noise, we randomly sample standard deviations from the range $\sigma \in [0.24, 11.51]$. For the signal-dependent noise, we randomly select from a set of camera NLFs. For the noise generated with Noise Flow, we feed the model with random camera identifiers and ISO levels. The $\sigma$ range, camera NLFs, ISO levels, and camera identifiers are all reported in the SIDD.
Furthermore, in addition to training with synthetic noise, we also train the DnCNN 
model with real noisy/clean image pairs from the SIDD-Medium and no noise augmentation (indicated as DnCNN-Real).

\begin{figure}[t]
    \centering
    \includegraphics[width=\columnwidth]{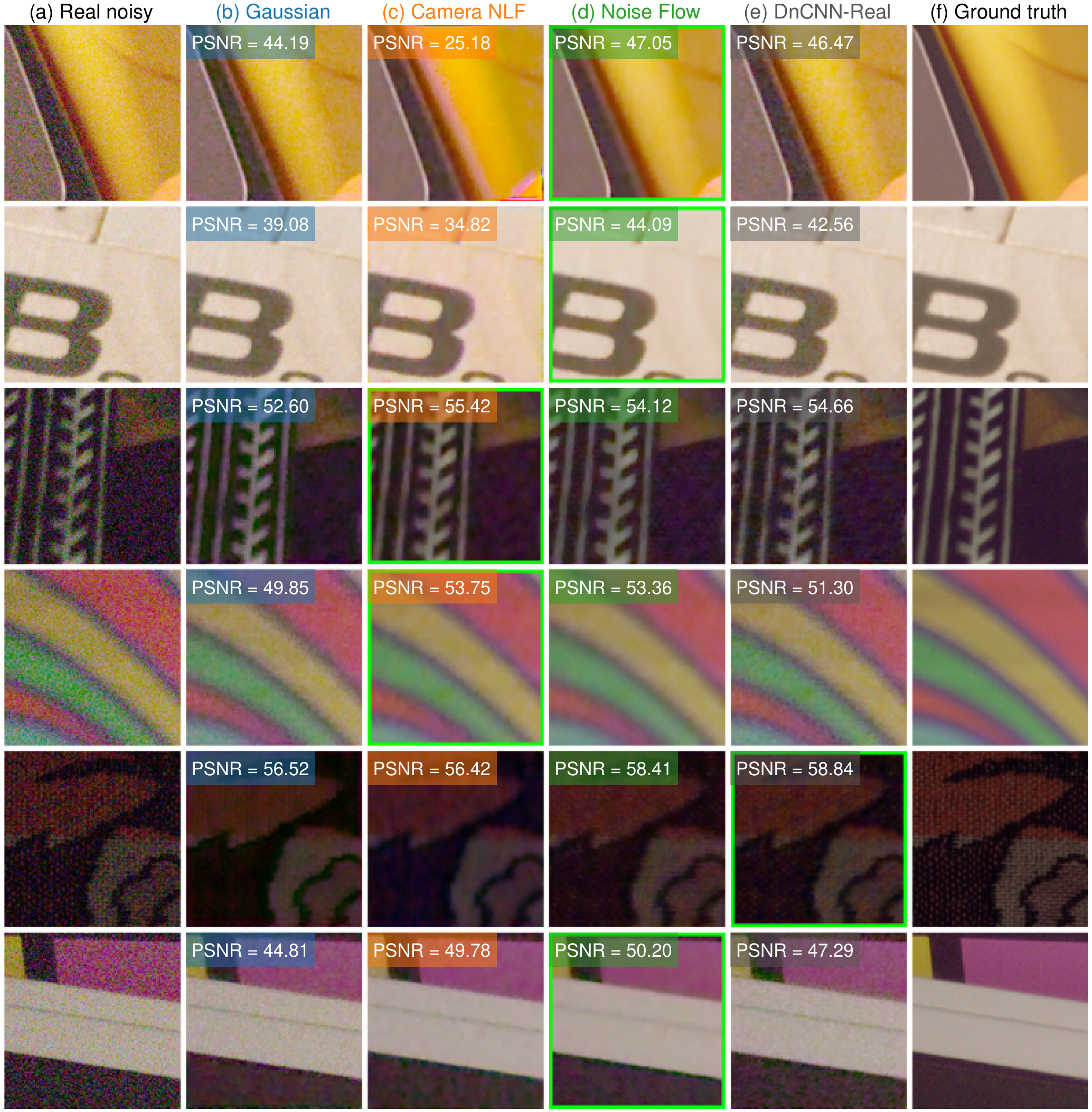}%
    \caption{Sample denoising results from DnCNN trained on three different noise synthesis methods: (b) Gaussian; (c) camera NLF; and (d) Noise Flow. (e) DnCNN trained on real noise. (a) Real noisy image. (f) Ground truth. 
    \vspace{-3mm}%
    \label{fig:denoising-samples}}
\end{figure}

\paragskip{Results and discussion} Table~\ref{tab:dncnn-results} shows the best achieved testing peak signal-to-noise ratio (PSNR) and structural similarity (SSIM)~\cite{Wang2004ImageSimilarity} of DnCNN using the aforementioned three noise synthesis strategies and the discriminative model trained on real noise. The model trained on noise generated from Noise Flow yields the highest PSNR and SSIM values, even slightly higher than DnCNN-Real due to the relatively limited number of samples in the training dataset.
We also report, in parentheses, the relative improvement introduced by DnCNN-NF over the other two models in terms of root-mean-square-error (RMSE) and structural dissimilarity (DSIMM)~\cite{oza2009StructuralVideos, Webb2003StatisticalRecognition}, for PSNR and SSIM, respectively. We preferred to report relative improvement in this way because PSNR and SSIM tend to saturate as errors get smaller; conversely, RMSE and DSSIM do not saturate.  
For visual inspection, in Figure~\ref{fig:denoising-samples}, we show some denoised images from the best trained model from the three cases, along with the corresponding noisy and clean images. DnCNN-Gauss tends to over-smooth noise, as in rows 3 and 5, while DnCNN-CamNLF frequently causes artifacts and pixel saturation, as in rows 1 and 5. Although DnCNN-NF does not consistently yield the highest PSNR, it is the most stable across all six images. 
Noise Flow can be used beyond image denoising in assisting computer vision tasks that require noise synthesis (\eg, robust image classification~\cite{Diamond2017DirtyData} and burst image deblurring~\cite{Aittala2018BurstNetworks}.
In addition, Noise Flow would give us virtually unlimited noise samples compared to the limited numbers in the datasets.


\begin{table}[t]
\centering
\resizebox{.9\columnwidth}{!}{
\begin{tabular}{lll}
\toprule
\multicolumn{1}{c}{Model} & \multicolumn{1}{c}{PSNR} & \multicolumn{1}{c}{SSIM} \\ \midrule
DnCNN-Gauss & $43.63$ ($43.0\%$) & $0.968$ ($75.6\%$)\\
DnCNN-CamNLF & $44.99$ ($33.4\%$) & $0.982$ ($56.0\%$)\\ 
DnCNN-NF & \aarnk{\num{48.52}} & \aarnk{\num{0.992}} \\ \hdashline
DnCNN-Real & $47.08$ ($15.3\%$) & $0.989$ ($27.5\%$) \\
\bottomrule
\end{tabular}%
} 
\caption{DnCNN denoiser~\protect\cite{Zhang2017BeyondDenoising} trained on synthetic noise generated with Noise Flow (DnCNN-NF) achieves higher PSNR and SSIM values compared to training on synthetic noise, from a Gaussian model or camera NLFs, and real noise. Relative improvements of DnCNN-NF over other models, in terms of RMSE and DSSIM, are in parentheses.%
 \label{tab:dncnn-results}%
 \vspace{-3mm}}%
\end{table}
%

\section{Conclusion}

In this paper, we have presented a conditional normalizing flow model for image noise modeling and synthesis that combines well-established noise models and the expressiveness of normalizing flows. As an outcome, we provide a compact noise model with fewer than 2500 parameters that can accurately model and generate realistic noise distributions with $0.42$ nats/pixel improvement (\ie, $52\%$ higher likelihood) over camera-calibrated noise level functions. We believe the proposed method and the provided model will be very useful for advancing many computer vision and image processing tasks. The code and pre-trained models are publicly available at: \url{https://github.com/BorealisAI/noise_flow}.

\section*{Acknowledgments}
This work was supported by Mitacs through the Mitacs Accelerate Program as part of an internship at Borealis AI.  This study was also funded in part by the Canada First Research Excellence Fund for the Vision: Science to Applications (VISTA) programme and an NSERC Discovery Grant.  Dr. Brown contributed to this article in his personal capacity as a professor at York University.  The views expressed are his own and do not necessarily represent the views of Samsung Research. Abdelrahman is partially supported by an AdeptMind scholarship.



{\small
\bibliographystyle{ieee_fullname}
\bibliography{egbib}
}

\end{document}